# STATISTICAL EVALUATION OF VISUAL QUALITY METRICS FOR IMAGE DENOISING

*Egiazarian K[1], Ponomarenko M[1], Lukin V[2], and Ieremeiev O[2]*

[1]Laboratory of Signal Processing, Tampere University of Technology, Finland
[2]Department of Signal Transmitting, Receiving and Processing, National Aerospace University, Ukraine

**ABSTRACT**

This paper studies the problem of full reference visual quality assessment of denoised images with a special emphasis on images with low contrast and noise-like texture. Denoising of such images together with noise removal often results in image details loss or smoothing. A new test image database, FLT, containing 75 noise-free 'reference' images and 300 filtered ('distorted') images is developed. Each reference image, corrupted by an additive white Gaussian noise, is denoised by the BM3D filter with four different values of threshold parameter (four levels of noise suppression). After carrying out a perceptual quality assessment of distorted images, the mean opinion scores (MOS) are obtained and compared with the values of known full reference quality metrics. As a result, the Spearman Rank Order Correlation Coefficient (SROCC) between PSNR values and MOS has a value close to zero, and SROCC between values of known full-reference image visual quality metrics and MOS does not exceed 0.82 (which is reached by a new visual quality metric proposed in this paper). The FLT dataset is more complex than earlier datasets used for assessment of visual quality for image denoising. Thus, it can be effectively used to design new image visual quality metrics for image denoising.

## 1. INTRODUCTION

PSNR is a full reference objective quality metric used in image processing applications since 1960s. During the last decades, there has been a notable growth in use of other image visual quality metrics [3-12], along with a development of test image databases for their verification [1, 2]. According to the recent perceptual quality evaluation on TID2013 database [1], the following metrics BMFF [4], PSNR-HA [5], FSIM [6] and SR-SIM [7] obtain the largest SROCC values to MOS. Note, however, that TID2013 database cannot provide an accurate verification of metrics for image denoising, since the PSNR values between denoised (distorted) images and the corresponding reference images in TID2013 differ by at least 3 dB [1]. Most of the existed quality metrics shall properly assess image visual quality for images with such a large difference in PSNRs. In [8], another test image database containing 'distorted' images filtered by various denoising methods was presented. Although the database in [8] is complex for visual quality assessment, most of metrics, including PSNR, provide acceptable correspondence to human visual system (HVS).

State-of-the-art methods of image denoising, including those based on deep convolutional neural network [13], become more and more sophisticated, however, the margin in PSNR values between best denoising methods does not exceed 0.5 dB. Considering a poor correspondence of PSNR to HVS, such a small difference in PSNR does not allow to determine which image denoising method is better.

For image denoising, image regions with low-contrast and noise-like texture are most complicated ones. Application of the most advanced image denoising methods to these images often results in detail loss and oversmoothing. As it was stated in [14], noise-like textures can visually mask the noise, thus, decreasing a level of noise suppression for such textures can result in a better visual appearance of the filtered image. Contrary to this, regular textures cannot effectively mask the noise due to ability of HVS to predict true values of pixels of such textures [14].

In the paper, we first develop a new test image database, called FLT, with different textures such as noise-like, low-contrast, and regular. FLT is build for the verification of full-reference image visual quality metrics. Next, we propose a modification of MSDDM metric [9], to provide a largest SROCC value among different quality metrics and MOS on FLT. It takes into account masking ability of non-predictable energy of image regions [14]. We have chosen MSDDM metric since this metric is based on dissimilarity maps and works with non-predictable energy of image regions.

The paper is organized as follows. In Section 2, the FLT database and experiments carried out to calculate MOS values, and their analysis, are described. A modification of the MSDDM metric is proposed in Section 3. A comparative analysis of well-known metrics using the proposed FLT database is carried out in Section 4.

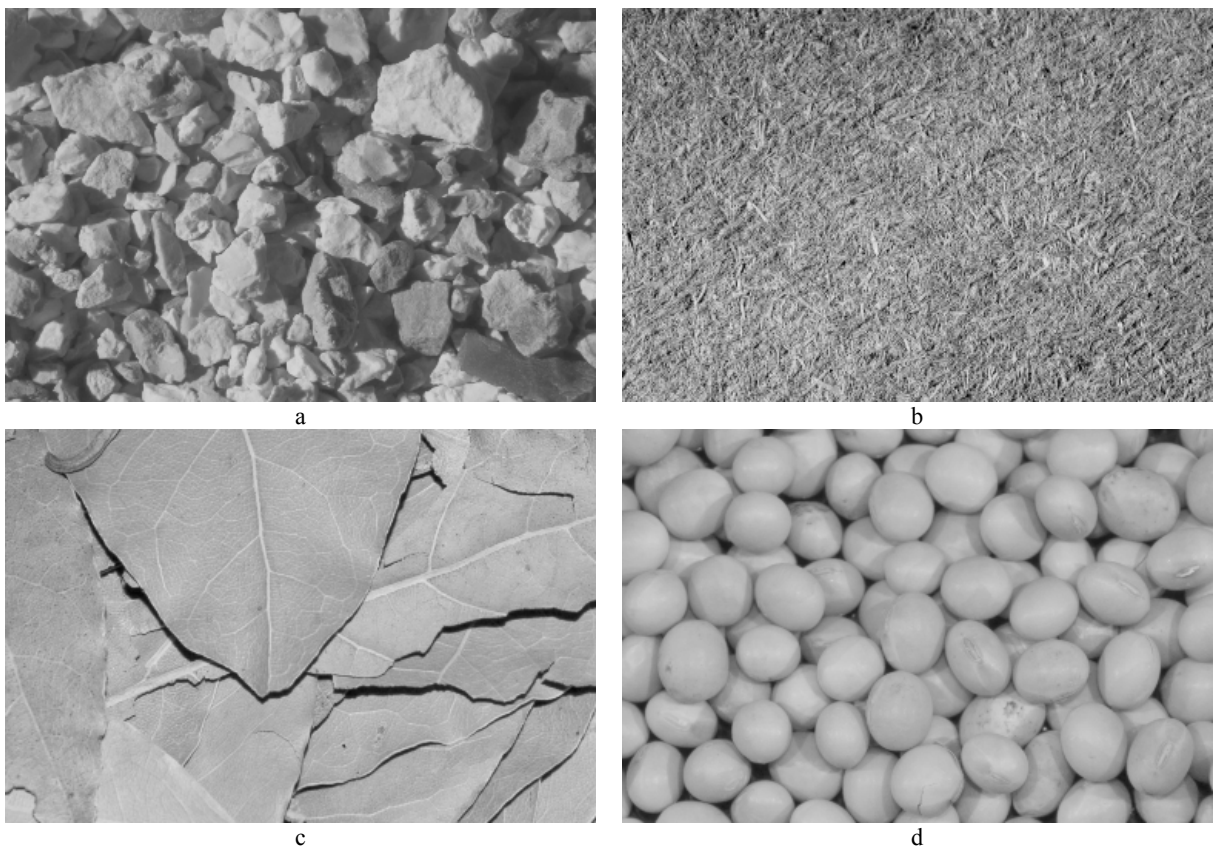

Figure 1. Examples of reference test images in FLT database

## 2. FLT TEST IMAGE DATABASE

The reference images for the developed FLT database are 75 grayscale texture images of resolution 384x256 pixels selected from Amsterdam Library of Textures [15]. Examples of these images are shown in Fig. 1. As one can see, FLT contains images with noise-like textures (Fig. 1, b) and images with regular textures (Fig. 1, d) which are well predictable for HVS. Part of images in FLT database are low-contrast textures (Fig. 1, c). Some of these images, like one presented in Fig. 1,a, can be placed somewhere in between noise-like and regular textures. Each test image has been distorted by an additive white Gaussian noise with variance $\sigma^2=200$. After this, distorted images have been filtered by BM3D filters [16] with four different thresholds (threshold values equal to $1.6\sigma$, $2.0\sigma$, $2.4\sigma$ and $2.8\sigma$). In this way, for each reference image, four distorted images with different levels of noise suppression have been obtained (larger threshold value corresponds to stronger noise suppression and, thus, stronger smoothing of textures).

To estimate visual quality of distorted images and obtain MOS values, 47 observers have been involved. A method of pairwise comparison of visual quality of distorted images, similar to the one in [1] has been employed (see Fig.2).

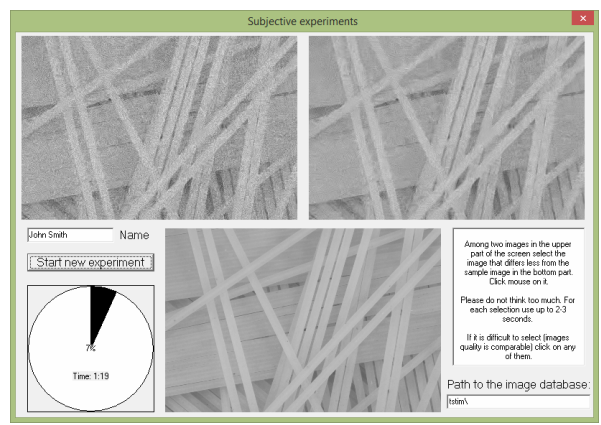

Fig 4. Screenshot of software used in experiments that illustrates positions of displayed images

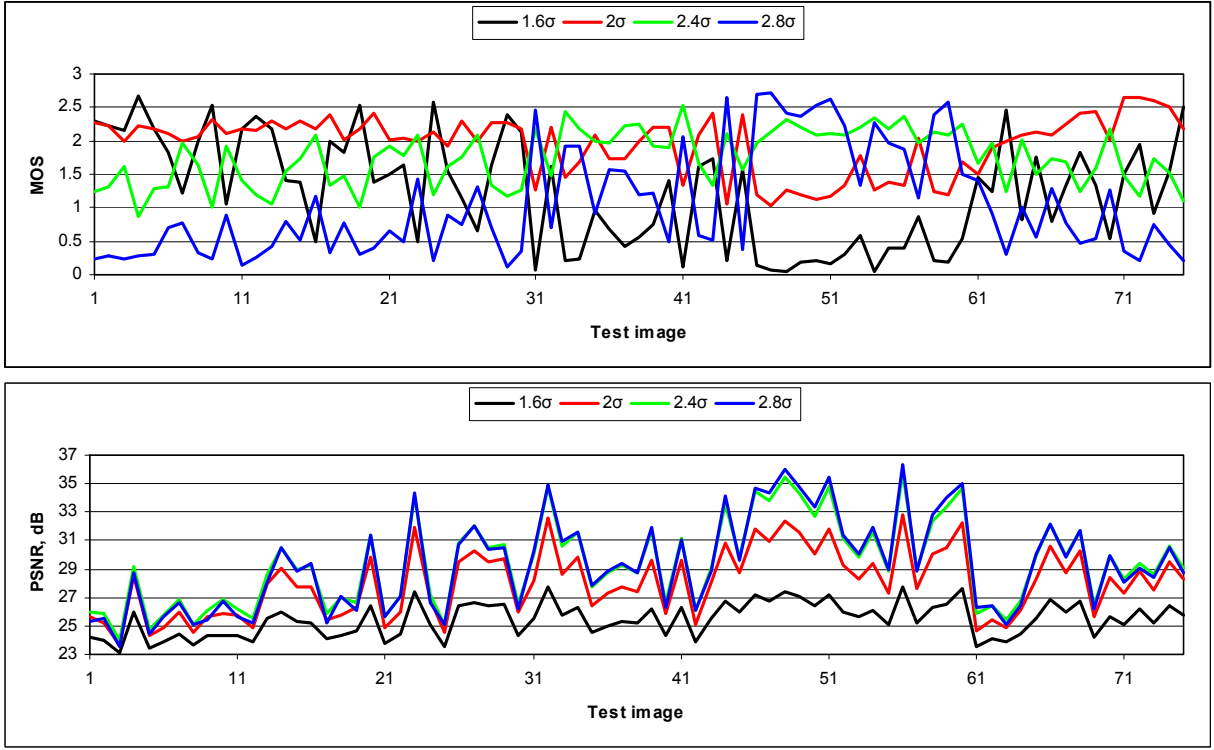

Fig.3. Values of MOS and PSNR for all test images of FLT database for four denoising levels

In correspondence to [17], a part of abnormal estimates (about 4%) was rejected before calculation of MOS.

The obtained values of MOS as well as values of output PSNR for each image of FLT database are shown in Fig. 3. Note that for BM3D filter the value of the threshold equal to 2.8σ is quasi-optimal according to PSNR metric, which is visible from Fig. 3. For most of images, PSNR indicates better quality of filtered images with a threshold value corresponding to the maximal considered level of noise suppression. However, for MOS (see Fig. 3), there is no such dependence. According to MOS, for different images, different thresholds (respectively, levels of noise suppression) provide best visual quality of filtered images. As a consequence, a value of SROCC [18] between MOS and PSNR for FLT database is 0.05 (which corresponds to practically no correlation between PSNR and MOS), and thus, inability of PSNR to choose a proper denoising setup.

## 3. PROPOSED MODIFICATION OF MSDDM METRIC

Here we would like to verify the hypothesis that a more unpredictable is a texture (or image region) the stronger is its ability to mask a noise [14]. We use MSDDM metric [9] as a basis, since it takes into account a dissimilarity of image regions.

In [9], MSDDM metric is defined as:

$$MSDDM = -\frac{1}{QW}\sum_{i=1}^{Q}\sum_{j=1}^{W}\left(\sqrt{D_{ij}} - \sqrt{D_{ij}^d}\right)^2, \quad (1)$$

where Q x W is an image size, $D_{ij}$ and $D_{ij}^d$ are dissimilarity maps [9] of reference and distorted images.

We modify (1) by decreasing each element of the sum on a value proportional to $\sqrt{D_{ij}^d}$ (masking effect of non-predictable energy of distorted image for the pixel (i,j)), resulting in the following DisSimilarity Index (DSI):

$$DSI = -\frac{1}{QW}\sum_{i=1}^{Q}\sum_{j=1}^{W}\max\left(0, \left|\sqrt{D_{ij}} - \sqrt{D_{ij}^d}\right| - \frac{\sqrt{D_{ij}^d}}{4.5}\right)^2, \quad (2)$$

where DSI is the proposed modification, 4.5 is the correcting factor obtained as a result of the optimization process.

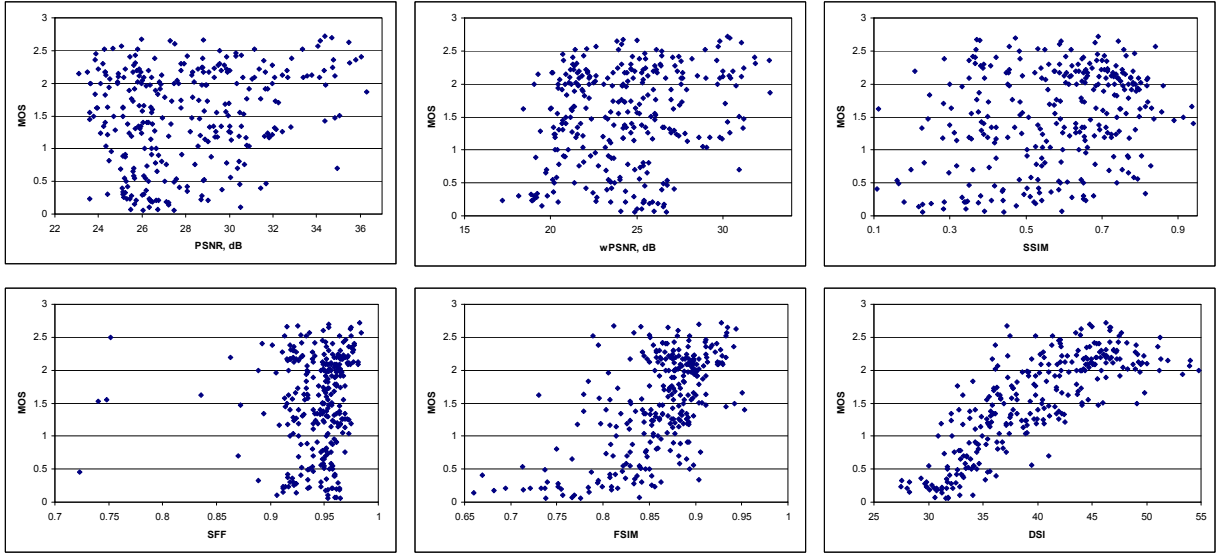

Fig. 4. Scatter-plots for various metrics and MOS for FLT database

In this paper, for calculation of the proposed DSI metric, we have used blocks of size 5x5 pixels as well as an area for blocks similarity calculation of 19x19 pixels.
A Matlab realization of the DSI metric can be found in [19].

## 4. COMPARATIVE ANALYSIS

In addition to the metrics PSNR, SSIM [2], MSDDM [9] and the proposed DSI, the metrics MSSIM [12], PSNR-HVSM [10], SR-SIM [7] as well as the metrics SFF [11] and FSIM [6] (among the best with respect to TID2013) are included in our comparative analysis. The metric wPSNR [8] is also added to the analysis since it is a simple alternative to PSNR, assigning a larger weight (set to 6 in this paper) to pixels which are distorted stronger due to filtering than corresponding noisy pixels.
We have calculated SROCC for all images of FLT database. Additionally, we split FLT database into three subsets. The first subset contains all images with noise-like textures (see an example in Fig.1, b). The second subset contains test images with low-contrast textures (see an example in Fig 1, c), and the third subset - images with regular textures (see an example in Fig 1, d).
In Table 1, values of SROCC for the compared metrics for all subsets of FLT database are presented. As one can see, for the majority of metrics, the most complicated is to assess visual quality of filtered images with low-contrast textures (subset #2).
It is also clear that the metrics PSNR and PSNR-HVSM are practically useless for estimation of visual quality of the filtered noise-like and low-contrast textures.

Table 1. SROCC values between MOS and the metrics

| # | Metric | Subset 1 (16 images) | Subset 2 (26 images) | Subset 3 (33 images) | Full set (75 images) |
|---|--------|------|------|------|------|
| 1 | DSI | 0.85 | 0.82 | 0.80 | 0.82 |
| 2 | wPSNR | 0.83 | 0.74 | 0.76 | 0.77 |
| 3 | MSDDM | 0.70 | 0.64 | 0.85 | 0.74 |
| 4 | FSIM | 0.75 | 0.58 | 0.79 | 0.71 |
| 5 | SR-SIM | 0.66 | 0.53 | 0.82 | 0.69 |
| 6 | SSIM | 0.58 | 0.32 | 0.73 | 0.55 |
| 7 | MSSIM | 0.58 | 0.21 | 0.72 | 0.51 |
| 8 | SFF | 0.21 | 0.22 | 0.69 | 0.42 |
| 9 | PSNR-HVSM | -0.17 | -0.12 | 0.72 | 0.23 |
| 10 | PSNR | -0.55 | -0.31 | 0.62 | 0.05 |

The proposed DSI metric for the database provides a correspondence to human perception better than others. However, the value of SROCC for this metric (0.82) in still low for its proper usage in estimation of effectiveness for image denoising.
Table 1 also shows that a simple metric wPSNR (with very low computational complexity) provides better correspondence to HVS than most of other metrics, and, thus, it can substitute PSNR and SSIM for visual quality assessment of denoised images.
Scatter-plots for some of considered metrics are shown in Fig. 5. It can be seen that SROCC value 0.55 for the SSIM metric has very low correspondence to human perception. Therefore, usage of such metrics as PSNR, SSIM, SFF, and PSNR-HVSM for image denoising to assess visual quality of filtered images cannot be justified.

## 5. CONCLUSIONS

In this paper, we propose a new test image database FLT with MOS, providing very effective tool for verification of image visual quality metrics intended for assessment of effectiveness of image denoising methods.

It is shown that none of the considered metrics provides larger than 0.9 SROCC value with human perception.

A new image visual quality metric DSI (modification of the MSDDM metric) is proposed. It is shown that DSI provides better correspondence to HVS for FLT database than any other considered metrics. DSI takes into account a masking effect of non-predictable energy of image regions, proving correctness of the hypothesis formulated in [14].

The FLT database and Matlab realization of DSI metric are available in [19].

## 6. ACKNOWLEDGMENT

This work is supported by Academy of Finland, project no. 287150, 2015-2019.

## 7. REFERENCES

[1] N. Ponomarenko, L. Jin, O. Ieremeiev, V. Lukin, K. Egiazarian, J. Astola, B. Vozel, K. Chehdi, M. Carli, F. Battisti, C.-C. Jay Kuo, "Image database TID2013: Peculiarities, results and perspectives", *Signal Processing: Image Communication*, vol. 30, 2015, pp. 55-77.

[2] Wang, Z., Bovik, A., Sheikh, H., Simoncelli, E.: Image quality assessment: from error visibility to structural similarity. *IEEE Transactions on Image Processing*, vol. 13, issue 4, pp. 600-612, New York, 2004.

[3] D. Chandler, "Seven Challenges in Image Quality Assessment: Past, Present and Future Research", *Signal Processing*, vol. 2913, 2013, pp. 1–53.

[4] Jin, L. Egiazarian, K. C-C. Jay Kuo. Performance comparison of decision fusion strategies in BMMF-based image quality assessment // Proceedings of *APSIPA*, Hollywood, 2012, pp. 1-4.

[5] Ponomarenko, N., Eremeev, O., Lukin, V., Egiazarian, K., Carli, M. Modified image visual quality metrics for contrast change and mean shift accounting // Proceedings of *CADSM*, pp. 305-311, 2011.

[6] L. Zhang, L. Zhang, X. Mou, and D. Zhang, "FSIM: a feature similarity index for image quality assessment", *IEEE Transactions on Image Processing*, vol. 20, No 5, pp. 2378-2386, 2011.

[7] Lin Zhang, Hongyu Li SR-SIM: A fast and high performance IQA index based on spectral residual // Proceedings of 19th *IEEE International Conference on Image Processing* (ICIP), 2012, pp. 1473 - 1476.

[8] N. Ponomarenko, S. Krivenko, K. Egiazarian, V. Lukin, J. Astola, "Weighted mean square error for estimation of visual quality of image denoising methods", *Proceedings of the international workshop on video processing and quality metrics, VPQM*, 2010, 6 p.

[9] Ponomarenko N., Jin L., Lukin V., Egiazarian K. Self-Similarity Measure for Assessment of Image Visual Quality // Proceedings of *13th International Conference Advances Concepts for Intelligent Vision Systems* (ACIVS), Ghent, Belgium, August 22-25, 2011, pp. 459-470.

[10] N. Ponomarenko, F. Silvestri, K. Egiazarian, M. Carli, J. Astola, V. Lukin, "On between-coefficient contrast masking of DCT basis functions", *Proceedings of the third international workshop on video processing and quality metrics, VPQM*, 2007, 4 p.

[11] Hua-wen Chang, Hua Yang, Yong Gan, and Ming-hui Wang Sparse Feature Fidelity for Perceptual Image Quality Assessment, *IEEE Transactions on Image Processing*, October 2013, vol. 22, no. 10, pp. 4007-4018.

[12] Wang Z., Simoncelli E.P. and Bovik A.C. Multi-scale structural similarity for image quality assessment, *IEEE Asilomar Conference on Signals, Systems and Computers*, 2003, pp. 1398-1402.

[13] Zhang, K., Zuo, W., Chen, Y., Meng, D., & Zhang, L., "Beyond a gaussian denoiser: Residual learning of deep cnn for image denoising", *IEEE Transactions on Image Processing.*, vol. 26, issue 7, pp. 3142 - 3155, 2017.

[14] O.I. Ieremeiev, N.N. Ponomarenko, V.V. Lukin, J.T. Astola, K.O. Egiazarian, "Masking effect of non-predictable energy of image regions", *Telecommunications and Radio Engineering*, 76 (8), pp. 685-708, 2017.

[15] G. J. Burghouts and J. M. Geusebroek, Material-specific adaptation of color invariant features, *Pattern Recognition Letters*, vol. 30, 2009, pp. 306-313.

[16] K. Dabov, A. Foi, V. Katkovnik, K. Egiazarian, "Image denoising by sparse 3-D transform-domain collaborative filtering", *IEEE Transactions on Image Processing*, 16(8), 2007, pp. 2080-2095.

[17] *ITU* (2002). Methodology for Subjective Assessment of the Quality of Television Pictures Recommendation BT.500-1, International Telecommunication Union, Geneva, Switzerland.

[18] Kendall M.G. *The advanced theory of statistics*. vol. 1, London, UK, Charles Griffin & Company limited, 1945, 457 p.

[19] *FLT and DSI* home page: http://ponomarenko.info/flt.htm